\documentclass[conference]{IEEEtran}
\IEEEoverridecommandlockouts
 \usepackage{nopageno}
\usepackage{geometry}
\usepackage{longtable}
\usepackage{amsmath,amssymb,amsfonts}
\usepackage{comment}
\usepackage{algorithmic}
\usepackage[backend=biber, style=numeric, sorting=none]{biblatex}
\usepackage{graphicx}
\usepackage{multirow}
\usepackage{subcaption}
\usepackage[switch]{lineno}
\usepackage{textcomp}
\usepackage{xcolor}
\def\BibTeX{{\rm B\kern-.05em{\sc i\kern-.025em b}\kern-.08em
    T\kern-.1667em\lower.7ex\hbox{E}\kern-.125emX}}

\makeatletter
\newcommand{\linebreakand}{%
  \end{@IEEEauthorhalign}
  \hfill\mbox{}\par
  \mbox{}\hfill\begin{@IEEEauthorhalign}
}
\makeatother

\newcommand{\setfirstpage}{
    \newgeometry{a4paper, top=70pt, left=54pt, right=54pt, bottom=54pt}
}
\newcommand{\setotherpages}{
    \newgeometry{a4paper, top=70pt, left=54pt, right=54pt, bottom=54pt}
}

\begin{document}

\ifnum\value{page}=1
    \setfirstpage
\else
    \setotherpages
\fi

\title{Knowledge Distillation Neural Network for Predicting Car-following Behaviour of Human-driven and Autonomous Vehicles\\
}

\author{
\IEEEauthorblockN{Ayobami Adewale}
\IEEEauthorblockA{\textit{Dept. of Civil and Env. Engineering} \\
\textit{University of Windsor}\\
Windsor, Canada \\
adewale2@uwindsor.ca}
\and
\IEEEauthorblockN{Chris Lee}
\IEEEauthorblockA{\textit{Dept. of Civil and Env. Engineering} \\
\textit{University of Windsor}\\
Windsor, Canada \\
cclee@uwindsor.ca}
\and
\IEEEauthorblockN{Amnir Hadachi}
\IEEEauthorblockA{\textit{ITS Lab, Institute of Computer Science} \\
\textit{University of Tartu}\\
Tartu, Estonia \\
hadachi@ut.ee}
\linebreakand
\IEEEauthorblockN{Nicolly Lima da Silva}
\IEEEauthorblockA{\textit{Dept. of Civil and Env. Engineering} \\
\textit{University of Windsor}\\
Windsor, Canada \\
ndasilva@uwindsor.ca}
}

\maketitle
\begin{abstract}
As we move towards a mixed-traffic scenario of Autonomous vehicles (AVs) and Human-driven vehicles (HDVs), understanding the car-following behaviour is important to improve traffic efficiency and road safety. Using a real-world trajectory dataset, this study uses descriptive and statistical analysis to investigate the car-following behaviours of three vehicle pairs: HDV-AV, AV-HDV and HDV-HDV in mixed traffic. The ANOVA test showed that car-following behaviours across different vehicle pairs are statistically significant (p-value $<$ 0.05).

We also introduce a data-driven Knowledge Distillation Neural Network (KDNN) model for predicting car-following behaviour in terms of speed. The KDNN model demonstrates comparable predictive accuracy to its teacher network, a Long Short-Term Memory (LSTM) network, and outperforms both the standalone student network, a Multilayer Perceptron (MLP), and traditional physics-based models like the Gipps model. Notably, the KDNN model better prevents collisions, measured by minimum Time-to-Collision (TTC), and operates with lower computational power, making it ideal for AVs or driving simulators requiring efficient computing.
\end{abstract}

\begin{IEEEkeywords}
 autonomous vehicles, human-driven vehicles, the Internet of Things, neural network knowledge distillation networks, and car-following behaviour.
\end{IEEEkeywords}

\section{Introduction}
In recent years, data-driven car-following models have been developed to analyze and predict car-following behaviours. Data-driven car-following models adopt advanced machine learning (ML) models such as 
Multiple-Layer-Perceptron (MLP) \cite{AdewaleLee2024}, Long Short-Term Memory (LSTM) networks \cite{LSTM2022}, graph neural network (GRNN) \cite{XingLiu2022}, and hybrid networks \cite{Qin2023CNNLSTM}.

These data-driven models have increasingly become the preferred method for predicting car-following behaviour, outperforming traditional physics-based models such as the Gipps model \cite{GIPPS1981105}, the Intelligent Driver Model (IDM) \cite{IDM} and the General Motors (GM) model \cite{CHAKROBORTY1999209}. While these classical physics-based models have undergone various improvements \cite{Shah2023, Hossain2021ImprovedCM} to increase their predictive capabilities, they still generally do not achieve the level of accuracy provided by data-driven models.

Despite the evident advantages of data-driven models in accuracy and adaptability, they often demand high computational power and memory resources, posing challenges for deployment in resource-constrained environments like IoT-based systems \cite{Zhang2020DeepLearningEdge, Liu2021EfficientNN}. While lightweight models like LightGBM offer efficiency \cite{9882329}, they struggle to capture the complex relationships in mixed traffic involving HDVs and AVs. Based on these limitations, there is a need for efficient model architectures to deliver high-performance predictions within the constraints of current IoT devices used in automotive technologies.

In this regard, this study introduces a Knowledge Distillation Neural Network (KDNN), which maintains a high level of predictive accuracy of car-following behaviour while considering the practical constraints of computational power and complexity. The KDNN distils complex and nuanced knowledge captured by large and complex neural network models into lightweight models suitable for deployment on devices with limited computational power and memory. It is expected that the KDNN can provide accurate predictions of car-following behaviour more efficiently with reduced computational time.

Thus, the objective of this study is to develop the KDNN for predicting car-following behaviour and evaluate its performance in comparison with the two conventional ML models—LSTM and Multiple Layer Perception Network (MLP). The models were trained and validated using real-world vehicle trajectories of autonomous vehicles (AV) and human-driven vehicles (HDV).
The remainder of the paper is organized as follows: Section II presents the dataset and analyzes car-following behaviour in AV-HDV mixed traffic using the data. Section III describes the KDNN. Section IV presents and discusses the results. Finally, Section V draws conclusions based on the findings.

\section{Data}
\subsection{Description}
This study analysed car-following behaviour using the Waymo trajectory dataset provided by Google, with processing detailed by \textcite{HU2022103490}. The dataset includes trajectories of HDVs and AVs equipped with Lidar and camera technologies. This dataset comprises various pairings of vehicles: 1032 pairs of HDV following HDV (HDV-HDV), 274 pairs of HDV following AV (HDV-AV), and 196 pairs of AV following HDV (AV-HDV).  Unfortunately, the dataset does not include instances of AVs following other AVs. Each vehicle's trajectory was recorded at a 0.1-second interval. 

Key variables collected from the dataset include the acceleration (\(a_n\), \(a_{n+1}\)) and jerk (\(j_n\), \(j_{n+1}\)) of both the lead vehicle (\(n\)) and the following vehicle (\(n+1\)), the speed (\(v_n\), \(v_{n+1}\)) of each vehicle, the spacing between them (\(x_n\)), and the speed difference between the following and lead vehicles (\(\Delta v_{n+1} = v_{n+1} - v_{n}\)).

\subsection{Descriptive Analysis of Car-following Behaviour}
To examine car-following behaviour in mixed traffic scenarios using the trajectory dataset, we focused on trajectory points where the distance between the following vehicle and the lead vehicle was under 50 meters for all three vehicle pairings. We also examined the speed variability among all three vehicle pairs for different bins of spacing. By looking at the speed variability, we can compare the inconsistency of vehicle speed for each vehicle pair. 

\begin{figure}[ht!]
\begin{center}
\includegraphics[width=0.90\linewidth]{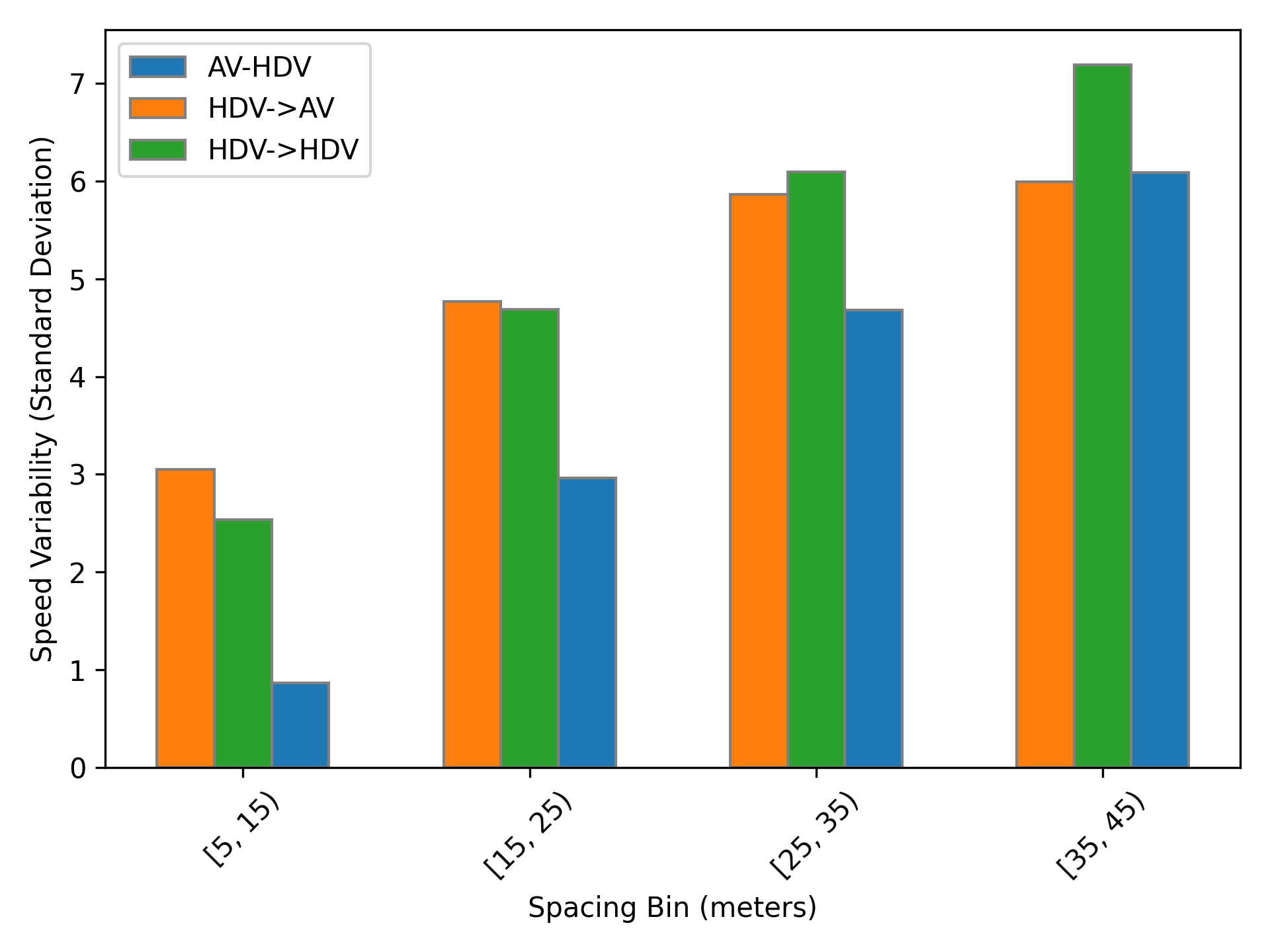}
    \caption{\textbf{Speed Variability}}
    \label{fig:speed_variability}
\end{center}
\end{figure}

Figure \ref{fig:speed_variability} shows that all three pairings exhibited increasing speed variability as spacing grew. The AV-HDV pair showed the lowest speed variability in tighter spacing bins ([5, 15] and [15, 25] meters), suggesting a more consistent and cautious approach. In contrast, HDV-AV had the highest variability, possibly due to more aggressive or unpredictable speed adjustments. At the same time, HDV-HDV pairs demonstrated the broadest range of variability, reflecting diverse human driving behaviours.

To further understand car-following behaviour, skewness and kurtosis were analyzed to evaluate driving aggressiveness and skill levels. Both methods have been previously used to assess driver aggressiveness and skill levels, as pointed out by \textcite{Chen2019DrivingBehaviors}. Skewness measures how much a real-valued random variable's distribution leans away from the mean. In contrast, Kurtosis measures the "tailedness" of the probability distribution of a real-valued random variable,

Higher skewness suggests more abrupt acceleration or deceleration, which reflects aggressive driving behaviour. Conversely, higher kurtosis suggests a driver's enhanced skill level, implying more consistent and controlled driving. 

\begin{figure*}[ht!]
\centering
    \subfloat[\centering Skewness of Speed Difference]{\includegraphics[width=0.40\textwidth]{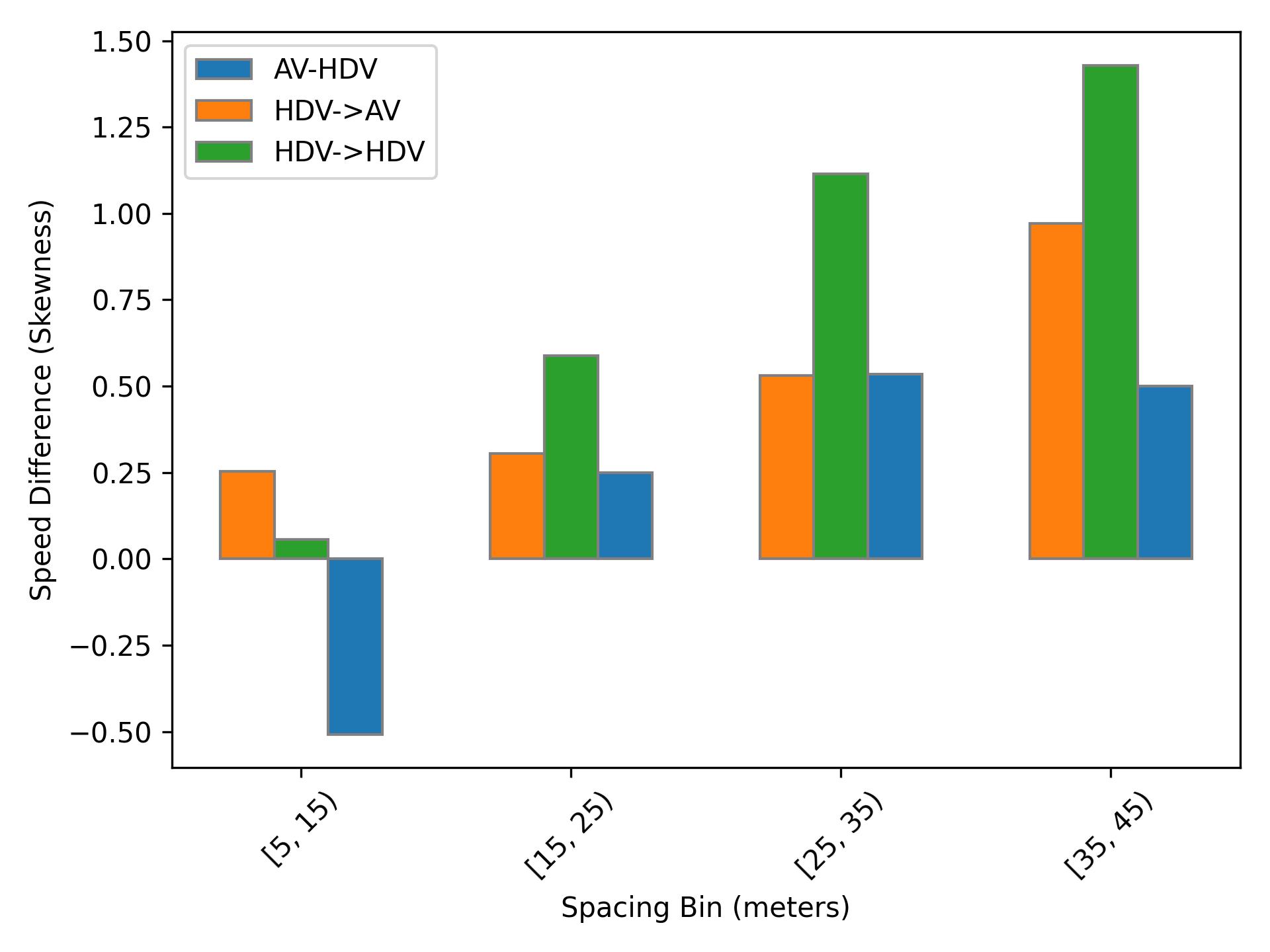}}\quad
    \subfloat[\centering Kurtosis of Speed Difference]{\includegraphics[width=0.40\textwidth]{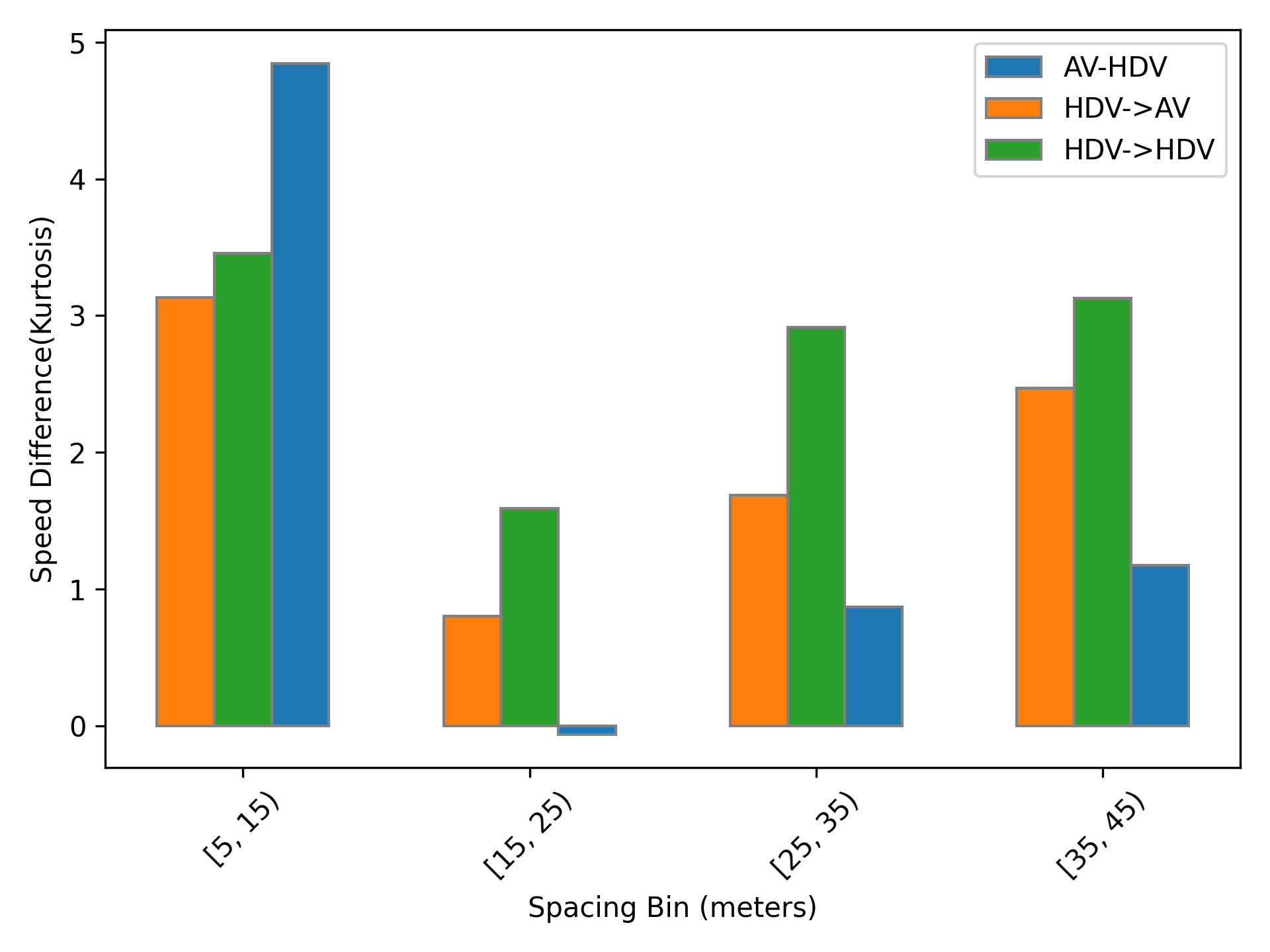}}
    \caption{\textbf{Skewness and Kurtosis}}
    \label{fig:statistics}
\end{figure*}

Figure \ref{fig:statistics} shows the Skewness and Kurtosis for all three vehicle pairs. From Figure \ref{fig:statistics}(a) we can see a pattern in the data, where aggressiveness generally increases with spacing. Notably, the AV-HDV pair had lower skewness, indicating steadier driving, while the HDV-HDV pair showed higher skewness, particularly in wider spacing bins. This suggests that human drivers are generally more aggressive. For kurtosis (Figure \ref{fig:statistics}(b)), AV-HDV pairs exhibited the highest values in close spacing, indicating precise speed adjustments by AVs, likely due to collision avoidance algorithms. HDV-AV and HDV-HDV pairs also showed high kurtosis but were less pronounced, suggesting that while human drivers can be skilful, they may not match the precision of AVs.

\subsection{Statistical Analysis of Car-following Behaviour}
To further analyse the effects of lead and following vehicle types on car-following behaviour, the means of the following vehicle's speed, acceleration, and TTC were compared among the three-vehicle pairs using a one-way ANOVA test. Given that these parameter values are sensitive to the spacing between the lead and the following vehicles, mean values were analyzed across three defined spacing categories: $<=10m$, $10m-15m$, and $15m-30m$. 

Table \ref{tab:combined_table} summarizes the statistical analysis results for the mean following speed, mean acceleration, and TTC for different vehicle pair types across the defined spacing categories.

The results show that mean speed increased with longer spacing for all vehicle pair types. The ANOVA test confirms that the mean speed differs significantly among the three vehicle pair types at a 95\ confidence interval (p-value $<0.05$). Notably, AV followers consistently displayed lower mean speeds than HDV followers, regardless of the lead vehicle type, likely due to AVs being programmed with a safety-first approach. Conversely, HDV drivers following AVs showed higher mean speeds than those following other HDVs across all spacing categories. This suggests that human drivers may trust and feel more comfortable following AVs.

Additionally, the table shows that mean acceleration generally increased with longer spacing, while mean deceleration was higher at shorter spacing. Like mean speed, mean acceleration also differed significantly among the vehicle pair types (p-value $<0.05$), indicating that vehicles tend to accelerate as spacing increases and decelerate as spacing decreases. Interestingly, HDV drivers following AVs exhibited significantly higher deceleration compared to those following other HDVs (p-value $<0.05$), consistent with findings that human drivers tend to decelerate more when following an AV due to the perceived predictability and safety of AVs \cite{Chen2022EffectsOA}.

Furthermore, for the spacing $<=10m$ and $10-15m$, the mean deceleration of AV-HDV was higher compared to HDV-HDV pairs. This is because AVs can better detect the spacing and apply brake earlier in short spacing to prevent a potential collision compared to HVs. However, for the spacing $15-30m$, the mean acceleration was positive, and the magnitude was similar for AV-HDV and HDV-HDV. This indicates that HDV drivers and AVs tend to accelerate when they perceive an opportunity to drive faster with longer spacing. Similarly, the mean deceleration was negative for the spacing $<=10m$, and the magnitude was similar for AV-HDV and HDV-HDV.

The analysis of TTC values showed that TTC was generally shorter at closer spacings. The mean TTC was significantly different among the vehicle pair types (p-value $<0.05$). For spacings $<=10m$ and $10m-15m$, AV-HDV pairs exhibited significantly higher TTC than HDV-HDV pairs (p-value $<0.05$), suggesting AVs can better control speed to maintain safer following distances. Although the mean TTC for AV-HDV pairs was slightly lower than for HDV-HDV pairs at $15m-30m$, the overall TTC values were still sufficiently high to prevent collision.

\begin{table*}[h!]
\scriptsize
\centering
\caption{\textbf{Statistical Analysis with ANOVA}}
\begin{tabular}{lccccccccc}
\hline
\multirow{2}{*}{Spacing Categories} & \multicolumn{3}{c}{Mean Following Speed (m/s)} & \multicolumn{3}{c}{Mean Acceleration (m/s\(^2\))} & \multicolumn{3}{c}{Time-to-Collision (seconds)} \\
\cline{2-10}
 & AV-HDV & HDV-AV & HDV-HDV & AV-HDV & HDV-AV & HDV-HDV & AV-HDV & HDV-AV & HDV-HDV \\
\hline
$\leq 10$m & 0.1 & 0.9 & 0.69 & -0.05 & -0.09 & -0.04 & 9.3 & 7.27 & 7.97 \\
10-15m & 0.95 & 5.27 & 3.98 & -0.07 & -0.04 & 0.07 & 10.4 & 8.73 & 9.77 \\
15-30m & 6.99 & 10.26 & 9.68 & 0.07 & -0.02 & 0.08 & 12.8 & 14.79 & 15.41 \\
\hline
\end{tabular}
\label{tab:combined_table}
\end{table*}

\section{Methods}
The KDNN is a machine learning approach where knowledge is transferred from a complex and large-scale model (teacher network) to a smaller and more efficient model (student network) \cite{stanton2021knowledge}. The KDNN mimics how human beings learn through the teacher-student relationship. This process is beneficial when deploying large models is impractical due to resource constraints.

Knowledge distillation works by training the student network to emulate the teacher network’s predictive behaviour. The teacher network, typically large and complex like LSTM or convolutional neural network(CNN), is first trained on the original dataset for high accuracy. The smaller and simpler student network learns from both the original data and the teacher's outputs, allowing it to learn complex patterns the teacher identifies.

The knowledge transfer between the teacher and student network occurs during the training phase of the network, and it can be done through three different approaches: 1) response-based transfer, 2) feature-based transfer, and 3) relation-based knowledge transfer \cite{gou2021knowledge}. This study uses response-based transfer, where the teacher network solely transfers its prediction to the student network. The KDNN is trained using a composite loss function, a weighted combination of the student network's predictions:

\begin{equation}
\mathcal{L}_{\text{MSE}}^{\text{Total}} = \alpha \mathcal{L}_{\text{MSE}}^{\text{Student}} + (1 - \alpha) \mathcal{L}_{\text{MSE}}^{\text{Distill}}
\end{equation}
where $\mathcal{L}_{\text{MSE}}^{\text{Total}}$ is the total loss, $\mathcal{L}_{\text{MSE}}^{\text{Student}}$ is the student network loss, $\mathcal{L}_{\text{Distill}}$ is the distillation loss, and $\alpha$ is a hyperparameter that manages the trade-off between the two.

Then, the total loss is re-written as follows:
\begin{equation}
\mathcal{L}_{\text{MSE}}^{\text{Total}} = \alpha \frac{1}{N} \sum_{i=1}^{N} (y^O_i - y^S_i)^2 + (1 - \alpha) \frac{1}{N} \sum_{i=1}^{N} (y^T_i - y^S_i)^2
\end{equation}

where $N$ represents the batch size, \(y^O_i\) is the observed value for the \(i\)-th instance, \(y^S_i\) is the student network's prediction for the \(i\)-th instance, \(y^T_i\) is the teacher network's prediction for the \(i\)-th instance.

Using the proposed KDNN, the speed of the following vehicle is predicted as follows:
\begin{equation}
    v_{n+1}(t+T) = f(S_{n+1}(t), v_{n}(t), \triangle v_{n+1}(t))
\end{equation}
where \(f(x)\) represents the KDNN function, \(T\) is the time-period of the input sequence, set to \(1.0\) seconds, and \(v_{n+1}(t+T)\) is the predicted speed at the next time step \(T\). \(S_{n+1}(t)\) is the distance between the lead and the following vehicles, \(v_{n}(t)\) is the speed of the lead vehicle, and \(\triangle v_{n+1}\) is the difference between the following and lead vehicle speeds. 

In this study, the Long-Short Term Memory (LSTM) model was used as a teacher network due to its ability to capture complex temporal relationships, as demonstrated in previous research. The Multiple Layer Perceptron (MLP) model was chosen as the student network for its simplicity and efficiency, which makes it more suitable for resource-constrained environments.

\section{Results and Discussion}
The KDNN model, LSTM model (teacher network), and MLP model (student network) were developed using Python in the Google Colab environment to leverage its higher processing power. Hyperparameter tuning was done using the `keras-tuner` library, with Randomized Cross-Validation (RandomCV) selected over an exhaustive search like GridSearch. Unlike GridSearch, RandomCV reduces computational demand by randomly sampling a subset of hyperparameter combinations when training the network, making it more efficient for complex models like LSTM.

To avoid overfitting or underfitting, TimeSeriesSplit cross-validation with KFold set to 3 was used. The dataset was split into 20\% for testing, with the remaining data divided into 20\% for validation and 60\% for training. The optimal configurations determined by RandomCV are as follows: The student network used 5 epochs, 2 hidden layers with 60 nodes each, ReLU for hidden layers, sigmoid for the output layer, a learning rate of 0.01, and a batch size of 100. The teacher network was configured with 10 epochs, 2 hidden layers (475 and 61 nodes), ReLU for hidden layers, sigmoid for the output layer, a dropout rate of 0.3, a learning rate of 0.0016, and a batch size of 161.

Table \ref{tab:alpha_distro} shows the root mean square errors (RMSE) for the teacher network, student network, and the KDNN model corresponding to different $\alpha$ values. Notably, the teacher network had lower  RMSE (0.585) than the student network (0.899). This shows that the teacher network has superior predictive capacity due to its complexity. The table also shows that the RMSE of the KDNN model varies with $\alpha$ in a non-linear relationship, and it was lowest when $\alpha$ = 0.5.
This RMSE was higher than the RMSE of the teacher network but lower than the RMSE for the student network. This indicates that the KDNN model successfully transferred knowledge from
the teacher network to the student network and balance the trade-off between prediction accuracy and computational efficiency. 

\begin{table}[ht!]
\scriptsize
\begin{center}
\caption{\textbf{Prediction error of KDNN model}}
\label{tab:alpha_distro}
\begin{tabular}{|ccc|}
\hline
\multicolumn{3}{c}{RMSE of teacher network = \textbf{0.585}} \\
\multicolumn{3}{c}{RMSE of student network = \textbf{0.899}} \\
\hline
Alpha $\alpha$ & RMSE & Error Difference with student network \\
\hline
0.1 & 1.394 & +0.495 \\
0.2 & 0.945 & +0.046 \\
0.3 & 0.777 & -0.122 \\
0.4 & 0.768 & -0.131 \\
\textbf{0.5} & \textbf{0.611} & \textbf{-0.288} \\
0.6 & 0.733 & -0.166 \\
0.7 & 0.950 & +0.051 \\
0.8 & 0.785 & -0.114 \\
0.9 & 0.991 & +0.092 \\
\hline
\end{tabular}
\end{center}
\end{table}

\begin{figure}[ht!]
\begin{center}
\includegraphics[width=0.80\linewidth]{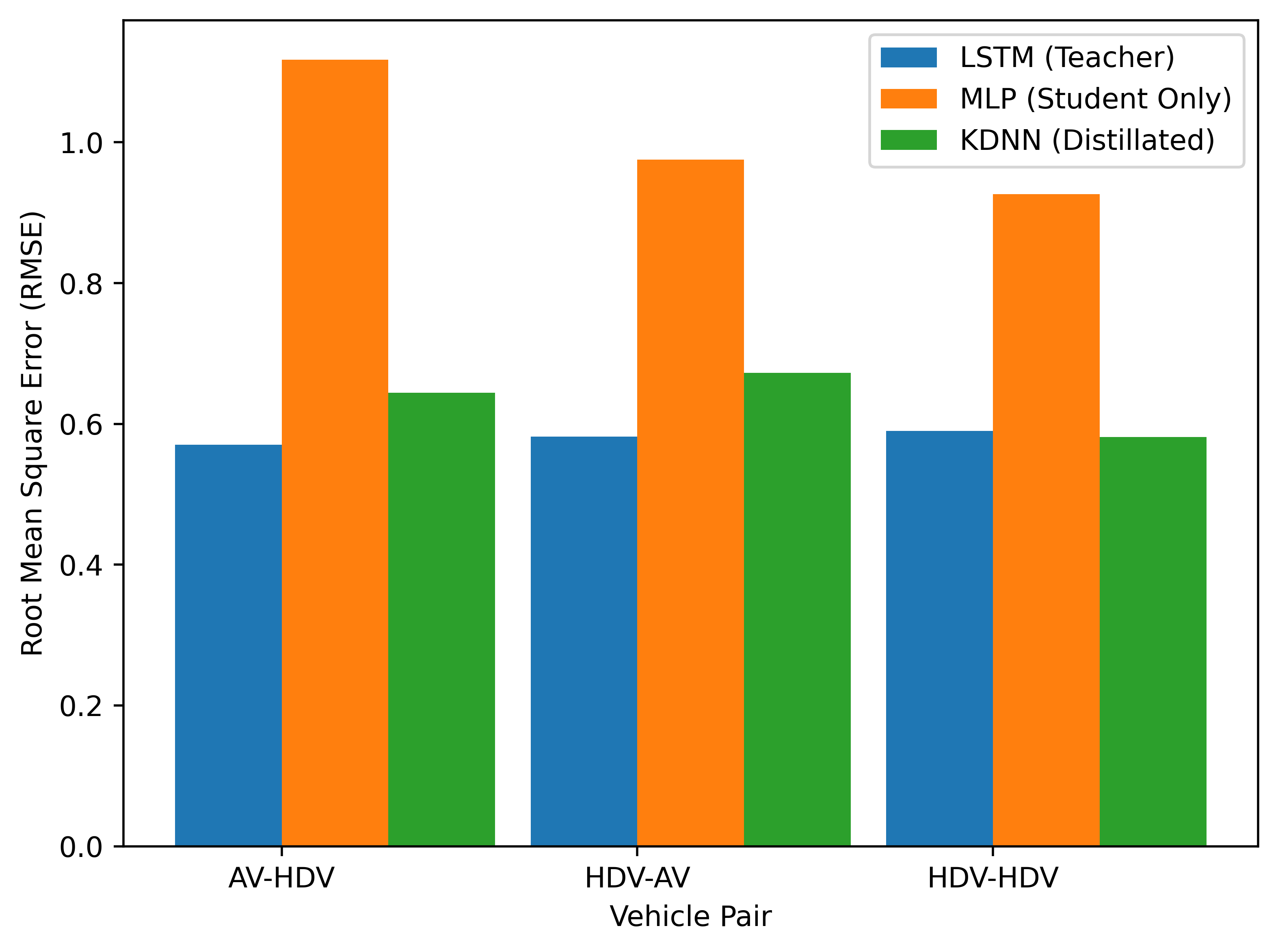}
    \caption{\textbf{Prediction errors by vehicle pair group}}
    \label{fig:prediction_error}
\end{center}
\end{figure}

Figure \mbox{\ref{fig:prediction_error}}  compares the prediction accuracy, measured by RMSE, among the teacher network, the student network, and the KDNN model by vehicle pair group. The LSTM teacher network consistently achieves the lowest RMSE values, showcasing its superior ability to capture complex temporal patterns in car-following behaviour. Conversely, the MLP student network has the highest RMSE values, particularly for the AV-HDV pair, indicating that its simpler structure struggles with capturing the complex interactions between AVs and HDVs. This may be due to the unpredictable nature of human decision-making and the algorithm-driven behaviour of AVs, which is challenging for a basic MLP network to model effectively.

The KDNN model significantly improves over the standalone MLP, with RMSE values that are particularly close to those of the LSTM in the HDV-HDV and AV-HDV pairs. This suggests that the knowledge distillation process effectively transfers essential predictive insights from the teacher network to the KDNN, making it more capable of handling the complexities of these interactions, especially where driving behaviour is more complex for a simpler network.

\begin{figure}[ht!]
\begin{center}
\includegraphics[width=0.80\linewidth]{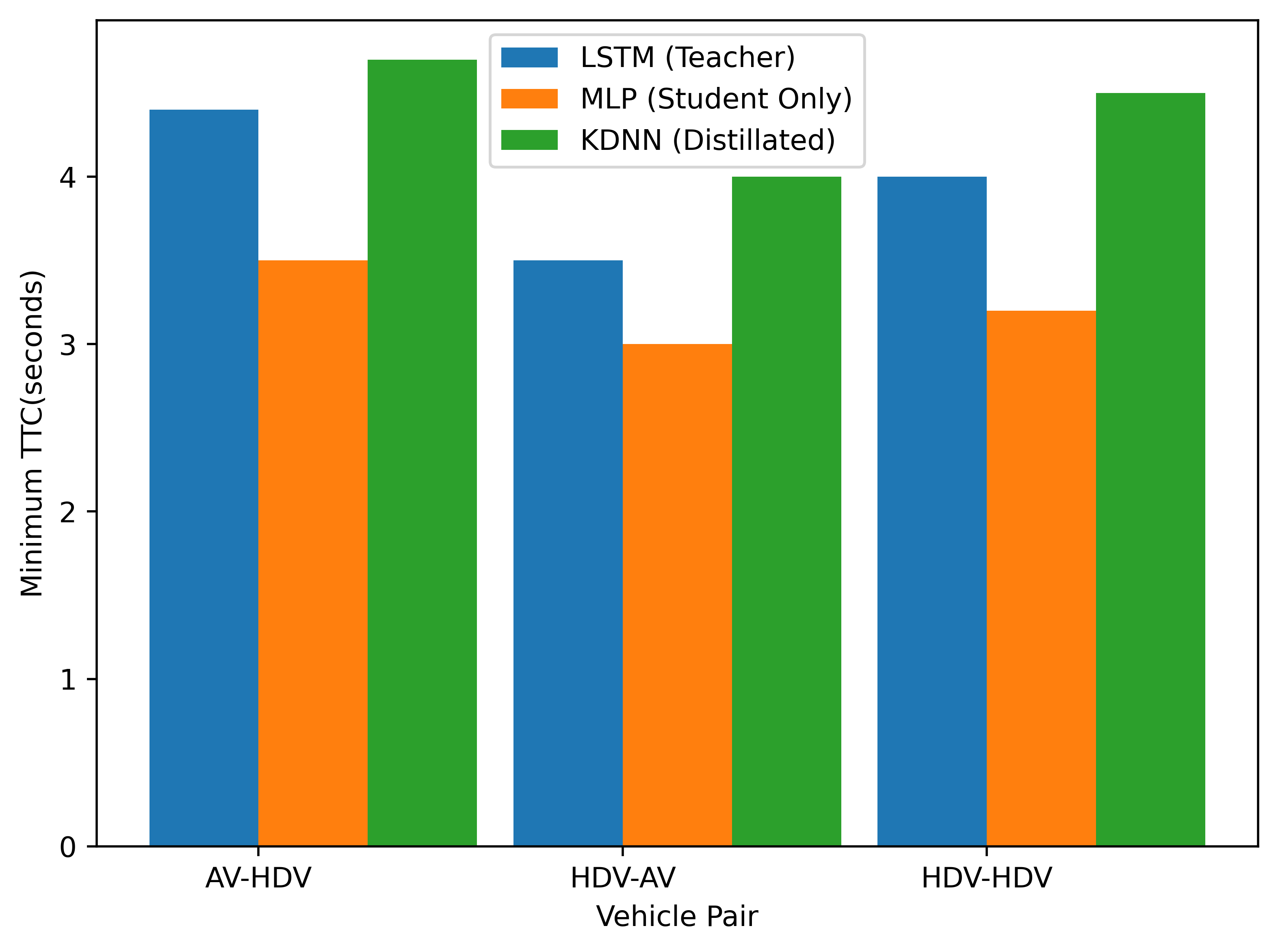}
    \caption{\textbf{Minimum time-to-collision by vehicle pair group}}
    \label{fig:min_ttc}
\end{center}
\end{figure}

The minimum Time-to-Collision (TTC) was compared among the three models to evaluate safety. Figure \ref{fig:min_ttc} shows that the KDNN model consistently had the highest minimum TTC, indicating safer car-following behaviour than the teacher (LSTM) and student (MLP) networks. This result is consistent with the findings by \textcite{9857598}.

\begin{figure}[ht!]
\begin{center}
\includegraphics[width=0.80\linewidth]{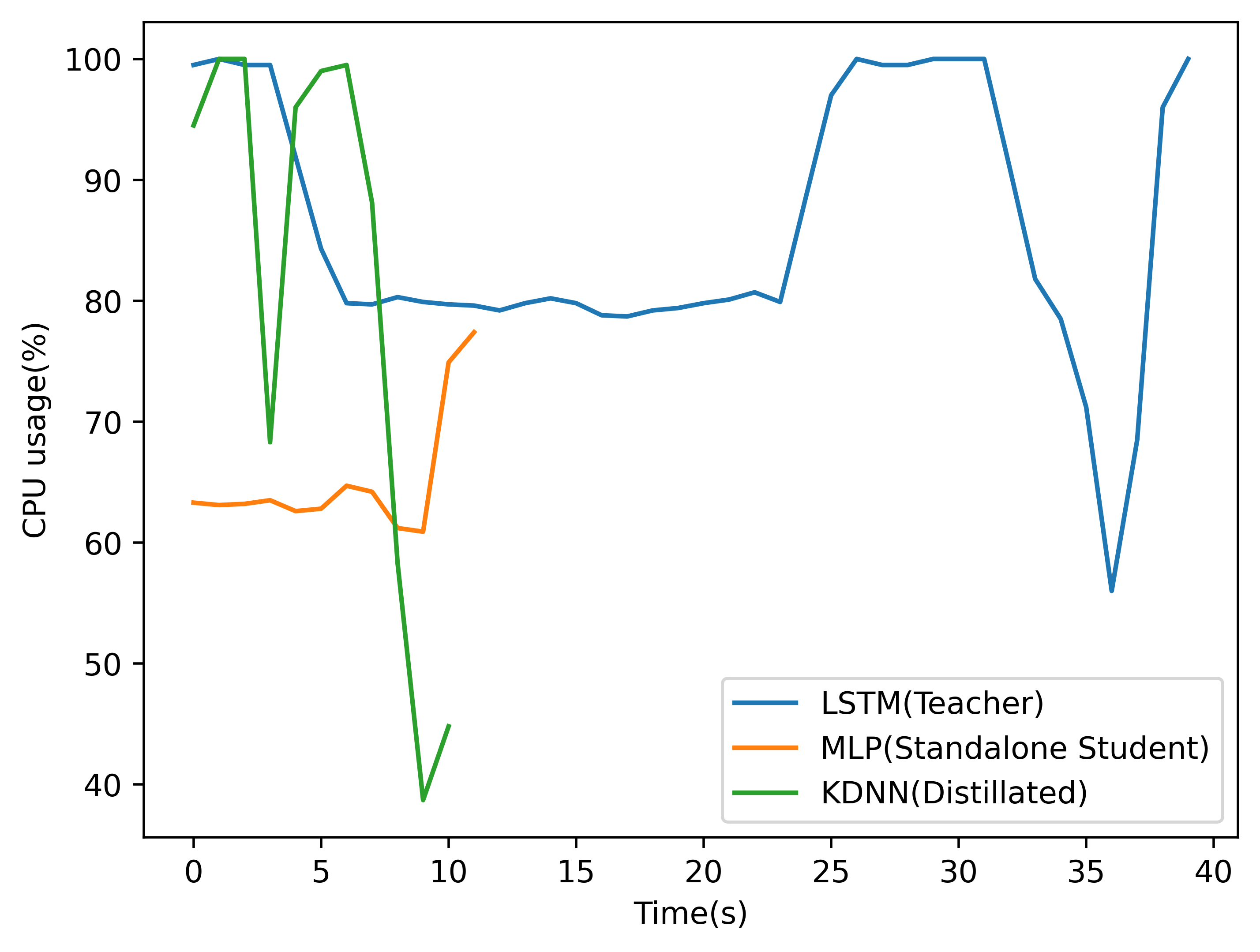}
    \caption{\textbf{CPU usage during testing phase}}
    \label{fig:cpu_usage}
\end{center}
\end{figure}

The CPU usage during their testing phase was compared among the three models as shown in Figure~\ref{fig:cpu_usage} to evaluate computational efficiency. The teacher network showed the highest CPU usage due to its complexity, while the student network had the lowest, reflecting its simpler structure. The KDNN model's CPU usage was between the two, demonstrating that it achieves a balance by maintaining higher accuracy without significantly increasing computational demand.

\begin{figure}[h!]
    \centering
    \subfloat[\centering Vehicle pair 1]{\includegraphics[width=0.30\textwidth]{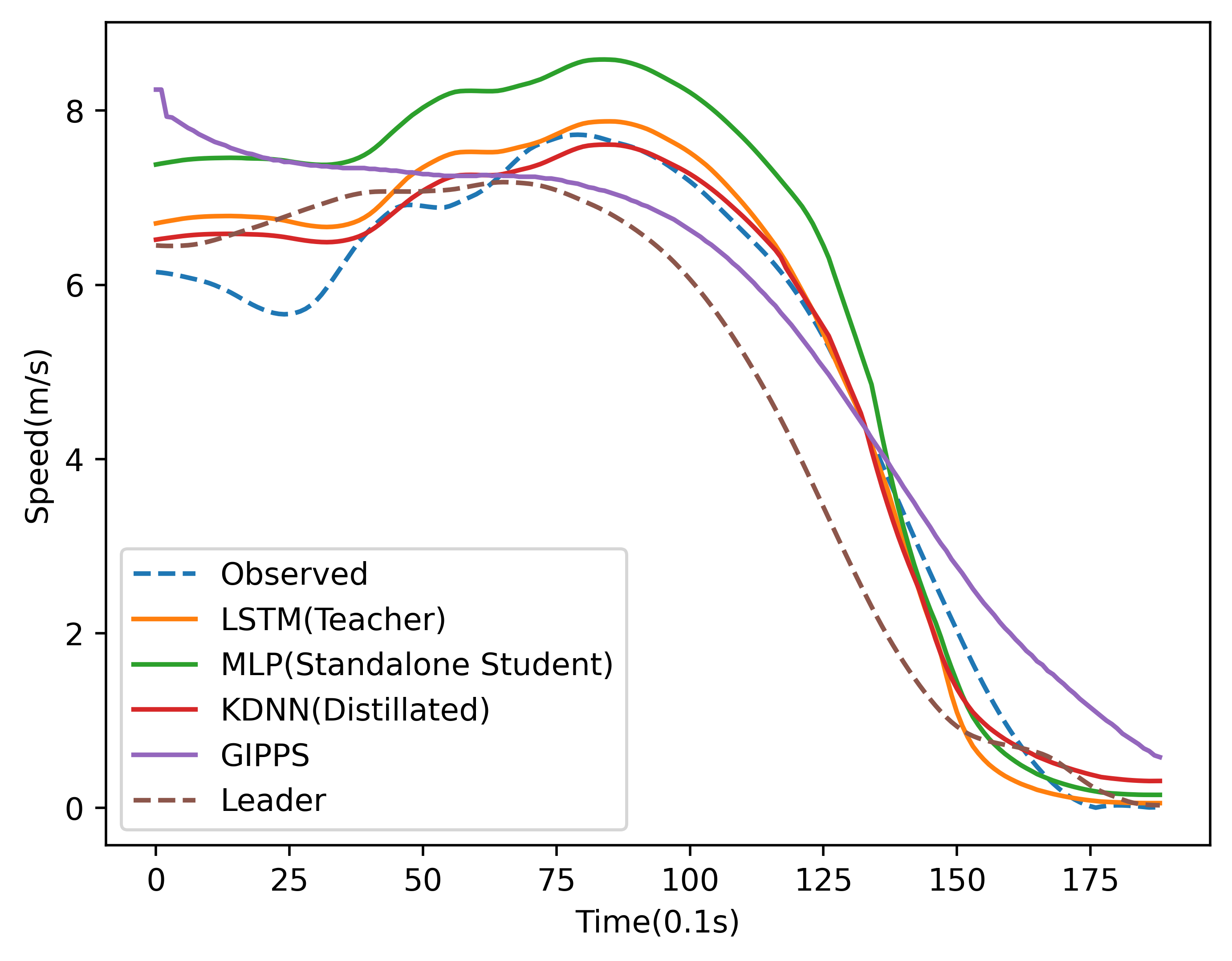}}\\
    \subfloat[\centering Vehicle pair 2]{\includegraphics[width=0.30\textwidth]{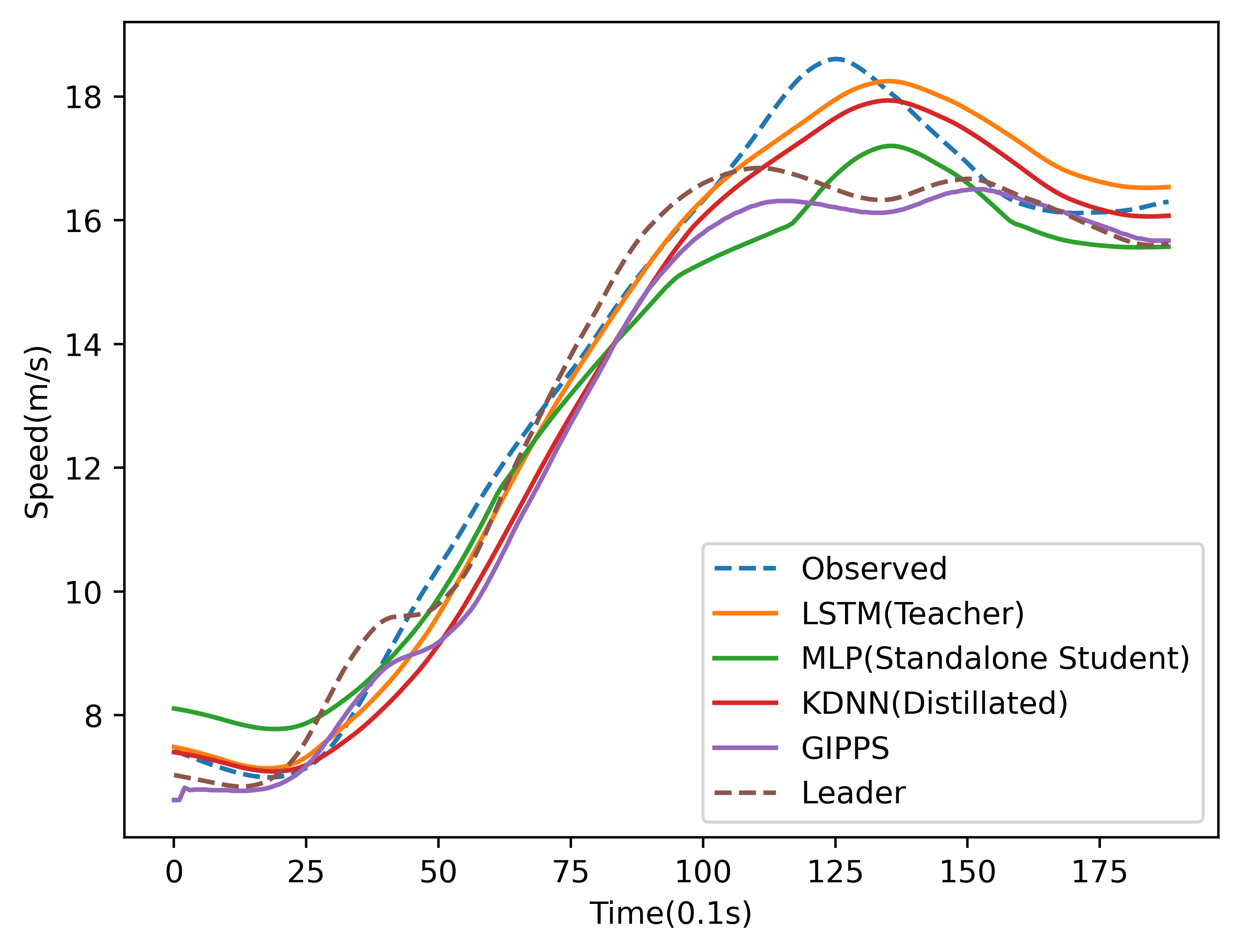}}
    \caption{\textbf{Observed and predicted following vehicle speed for teacher network, student network, and KDNN model}}
    \label{fig:lstm-pt-gipps}
\end{figure}

The predicted and observed speed profiles were compared among models for six vehicle pairs using four models: the teacher network (LSTM), student network (MLP), KDNN, and Gipps model. Table \ref{tab:rmse-all} shows the RMSE values, with the KDNN model generally matching or outperforming the others, particularly in complex scenarios. Figure \ref{fig:lstm-pt-gipps} shows that the KDNN model's predictions closely align with those of the teacher network, indicating that it effectively captures the underlying patterns learned through the teacher network.

\begin{table}[ht!]
\scriptsize
\begin{center}
\caption{\textbf{Prediction errors for 6 vehicle pair }}
 \begin{tabular}{ |p{0.6cm} |p{1.5cm} |p{1cm}| p{1.1cm}| p{0.8cm}| p{0.8cm} |} 
 \hline
 Pair & Pair group & GIPPS & MLP & LSTM  & KDNN\\
 \hline
1 & AV - HDV & 0.985 &  0.842 & 0.819 & 1.258\\
 \hline
2 & AV - HDV & 1.845 &  1.086 & 0.469 &  0.303 \\
 \hline
3 & HDV - AV & 0.883 &  0.984 & 0.242 & 0.137 \\
 \hline
4 & HDV - AV & 1.060 & 0.791 & 0.244 &  0.408 \\
 \hline
5 & HDV - HDV & 0.369 & 0.175 & 0.509 & 0.104 \\
 \hline
6 & HDV - HDV & 1.061 &  0.950 & 0.607 & 0.200 \\
 \hline
\end{tabular}
\label{tab:rmse-all}
\end{center}
\end{table}

\section{Conclusions and Recommendations}
This study analyzes car-following behaviour in mixed traffic of Human-driven Vehicles (HDVs) and Autonomous Vehicles (AVs) using real-world vehicle trajectory data. It also applies a knowledge distillation neural network (KDNN) model to predict car-following behaviour in mixed traffic. 

The analysis showed that AVs had higher mean Time-to-Collision (TTC) in closer spacing than HDVs, reflecting AVs' safety features. HDVs, however, exhibited higher mean speeds when following AVs, suggesting greater comfort in following AVs than other HDVs.

In comparing RMSE among different models, the KDNN model showed better predictive accuracy than the standalone MLP student network and higher computational efficiency than the LSTM teacher network, as indicated by lower CPU usage. Additionally, the KDNN model resulted in safer driving behaviour, evidenced by higher minimum TTC values across all vehicle pair groups. This suggests that the KDNN model can replicate more conservative and safer driving behaviour, which can be adapted to the AV control algorithm. 

In conclusion, the KDNN model offers a balanced solution with accurate predictions and computational efficiency, making it suitable for resource-constrained environments like edge computing in AVs. Future research should explore the transferability of the KDNN model using other trajectory datasets like NGSIM and UBER. Different machine-learning models for teacher and student networks should also be considered to improve the balance between prediction accuracy and computational power demand.
\printbibliography

\end{document}